\pdfoutput=1
\documentclass[11pt]{article}

\usepackage[final]{acl}
\usepackage{times}
\usepackage{latexsym}
\usepackage{amsmath}

\usepackage[T1]{fontenc}
\usepackage[utf8]{inputenc}

\usepackage{microtype}

\usepackage{inconsolata}

\usepackage{booktabs} % for professional tables
\usepackage{times}
\usepackage{wasysym}
\usepackage{xcolor}
\usepackage{soul}
\usepackage{siunitx}
\usepackage{tcolorbox}
\usepackage{subfig}
\usepackage{tikz}
\usepackage{listings}
\usepackage{color}
\usepackage{latexsym}
\usepackage{multirow}
\usepackage{colortbl}

\usepackage{graphicx}
\usepackage{hyperref}
\usepackage{multirow}
\usepackage{booktabs}
\usepackage{algorithm,algcompatible,amsmath}
\usepackage{mathtools}
\usepackage{tabularx}
\usepackage{xcolor}
\usepackage{colortbl}
\usepackage{tablefootnote}
\newcommand{\methodFull}{\textbf{Mu}lti-\textbf{D}ocument \textbf{A}ttention \textbf{F}ocusing}
\newcommand{\method}{\textit{MuDAF}}
\usepackage{enumitem}
\setitemize[1]{itemsep=0pt,partopsep=0pt,parsep=\parskip,topsep=5pt}
\title{MuDAF: Long-Context Multi-Document Attention Focusing through Contrastive Learning on Attention Heads}

\author{Weihao Liu\thanks{Work done during their internship at Microsoft.},\quad Ning Wu,\quad Shiping Yang$^{*}$,\quad Wenbiao Ding, \\ \bf{Shining Liang,\quad Ming Gong,\quad Dongmei Zhang} \\
Microsoft Corporation, Beijing, China \\ 
\texttt{liuweihao2022@outlook.com} \quad \texttt{\{wuning, v-shipiyang\}@microsoft.com} \\
\texttt{\{wenbiaoding, shiningliang, migon, dongmeiz\}@microsoft.com} 
}

\begin{document}
\maketitle
\begin{abstract}

Large Language Models (LLMs) frequently show distracted attention due to irrelevant information in the input, which severely impairs their long-context capabilities. Inspired by recent studies on the effectiveness of retrieval heads in long-context factutality, we aim at addressing this distraction issue through improving such retrieval heads directly. We propose~\methodFull~(\method), a novel method that explicitly optimizes the attention distribution at the head level through contrastive learning. According to the experimental results,~\method~can significantly improve the long-context question answering performance of LLMs, especially in multi-document question answering. Extensive evaluations on retrieval scores and attention visualizations show that~\method~possesses great potential in making attention heads more focused on relevant information and reducing attention distractions\footnote{\href{https://github.com/NeosKnight233/MuDAF}{https://github.com/NeosKnight233/MuDAF}}.

\end{abstract}

\section{Introduction}

As large language models (LLMs) continue to advance and find broader applications, the demand for their ability to efficiently handle ultra-long texts is growing. For instance, in Retrieval-Augmented Generation (RAG) systems~\citep{gao2023retrieval, jin2024long} and LLM agent systems~\citep{guo2024large}, models are often required to extract critical information from long-text corpora to accomplish complex generative tasks. However, research has shown that the real context window size of existing models often falls short of their claimed capabilities~\citep{an2024does}, revealing significant shortcomings in their ability to utilize information from long inputs. This issue is particularly evident in two major challenges: the "lost-in-the-middle"~\citep{liu-etal-2024-lost} phenomenon, where the middle portions of the text are neglected, and the interference from irrelevant information~\citep{shi2023large, wu2024how}. These challenges substantially hinder the performance of models in long-context tasks.

In recent years, many studies have conducted in-depth analyses of the role of attention mechanisms in long-context modeling~\citep{chen2024attention, hong2024token, zheng2024attention}. Notably, some research has identified that specific attention heads found in the \textbf{Needle-in-a-Haystack (NIAH)} test are critical for long-context factuality and has named them retrieval heads~\citep{wu2024retrieval}, which can perform a copy-paste operation from the input context to the output. Inspired by these works, we are extremely curious about such a question: \textit{How can we strengthen these retrieval heads directly to enhance models' long-context modeling capabilities?} 

\begin{figure}[t]
    \centering
    \includegraphics[width= 0.49\textwidth]{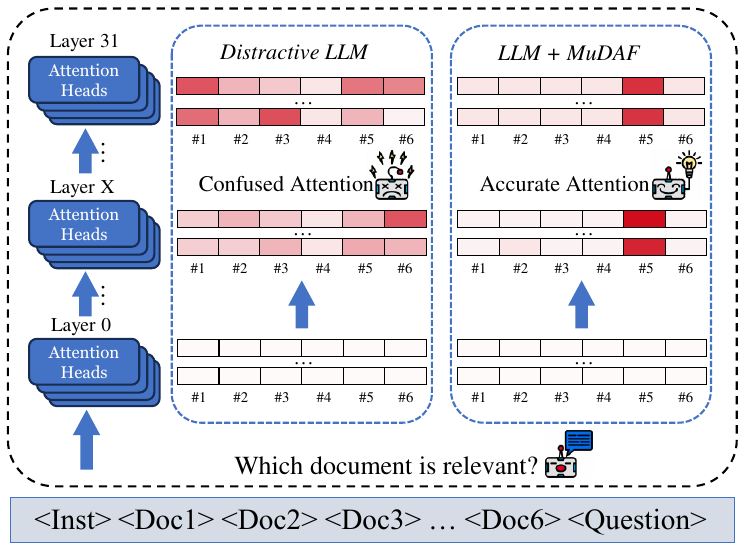}
    \caption{Given instructions, long documents and a specific question, LLMs can often be confused when facing information from multiple sources. Our method \textit{MuDAF} helps LLMs focus on documents related to the given question. Deeper colors represent higher attention values.}
    \label{fig:fig1}
\end{figure}

In this work, we care about \textbf{long-context question answering~(LCQA)}, especially \textbf{multi-document question answering~(MDQA)}, where the long input context contains many irrelevant documents and causes distractions. However, the retrieval heads in MDQA might be different from those in the NIAH test, since the NIAH test only shows the effectiveness of these heads in the copy-paste pattern. Considering the gap between them, we need to identify the retrieval heads in the MDQA setting and prove their effectiveness in helping models utilize relevant information in the context. As expected, we indeed found some retrieval heads that were different from those found in the NIAH test~(\textsection\ref{subsec:IR}), which may point to the fact that attention heads exhibit different levels of retrieval capabilities when applied to different tasks. Then, we explored methods to improve such retrieval heads in MDQA. As we know that attention weights are calculated by the softmax of the scaled dot product between query and key projections~\citep{vaswani2017attention}, it's feasible to optimize the attention weights allocation by learning better projections. Therefore, we propose \textbf{Mu}lti-\textbf{D}ocument \textbf{A}ttention \textbf{F}ocusing~(\textbf{\method}), a method that applies contrastive learning on attention heads to help them learn better query-key projections, thus optimizing their attention distributions. As depicted in Figure~\ref{fig:fig1}, \textit{MuDAF} can help attention heads be more focused on relevant passages while minimizing interference from irrelevant content.
%%%%%%%%%%%%%%%%%% need to be major revised %%%%%%%%%%%%%

In summary, our contributions are as follows.

\begin{itemize}
    \item  We provide a method to assess the retrieval capabilities of attention heads in multi-document question answering, distinguishing special retrieval heads that are different from those found in the NIAH test.
    \item We propose \method, a novel approach based on contrastive learning that optimizes the attention pattern at the head level to improve the long-context modeling ability of LLMs, especially in MDQA tasks.
    \item Experiments show that our methods can significantly enhance the long-context performance of LLMs and surpass GPT-4o in some datasets.
    \item  We did further analysis and ablations to show the effectiveness of our methods. Providing several insights about enhancing retrieval heads in MDQA.
\end{itemize}

\section{Related Work}

\paragraph{Attention-Based Salience for Long-Context.} Since the attention mechanism was first introduced by~\citep{bahdanau2014neural}, attention weight has become an important tool for interpreting important information in the input sequence~\citep{serrano-smith-2019-attention, ferrando2024primer}. For example, ~\citet{peysakhovich2023attention} use attention weights to estimate the importance of documents that can be leveraged to arrange their positions, thus improving the performance of long-context LLMs. \citet{xiao2024duoattention} manage to reduce KV cache for attention heads based on their attention patterns. \citet{he2024seekr} investigate the importance of attention weights in knowledge retention. Obviously, the attention mechanism has not only been a critical and reliable information resource for processing various long-context tasks~\cite{xiaoefficient, chen2024attention} but also presented substantial potential for further exploration and optimization~\citep{wu2024retrieval, lu2024longheads, he-etal-2024-never}. Our approach also highlights the function of certain attention heads in in-context retrieval~\citep{ram2023context}, aiming at optimizing attention distribution to get better long-context LLMs.

\paragraph{Distractions by Irrelevant Content.} Previous research has shown that LLMs can be easily disturbed by irrelevant context~\cite{shi2023large, wu2024easily}, making them overallocate attention to useless content. Some methods have been proposed to mitigate such issues.~\citet{liu2024bridging} introduce an innovative framework that helps LLMs recognize relevant entities in long contexts through efficient reference management.~\citet{wu2024reducing} reduce distractions by aligning the representations of the original context and the retrieved sub-context.~\citet{xiong2024artificial} enhance the retrieval capabilities of LLMs in highly similar contexts through fine-tuning on synthetic data. Another method proposes a novel differential attention mechanism to amplify attention to the relevant context while canceling attention noise~\citep{ye2024differential}. However, these methods do not explicitly optimize the attention distribution based on the input context, while our method provides a more straightforward and effective way.

%%%%%%%%%%%%%%%%%% need to be major revised %%%%%%%%%%%%%

\begin{figure*}[th]
    \centering
    \includegraphics[width= \textwidth]{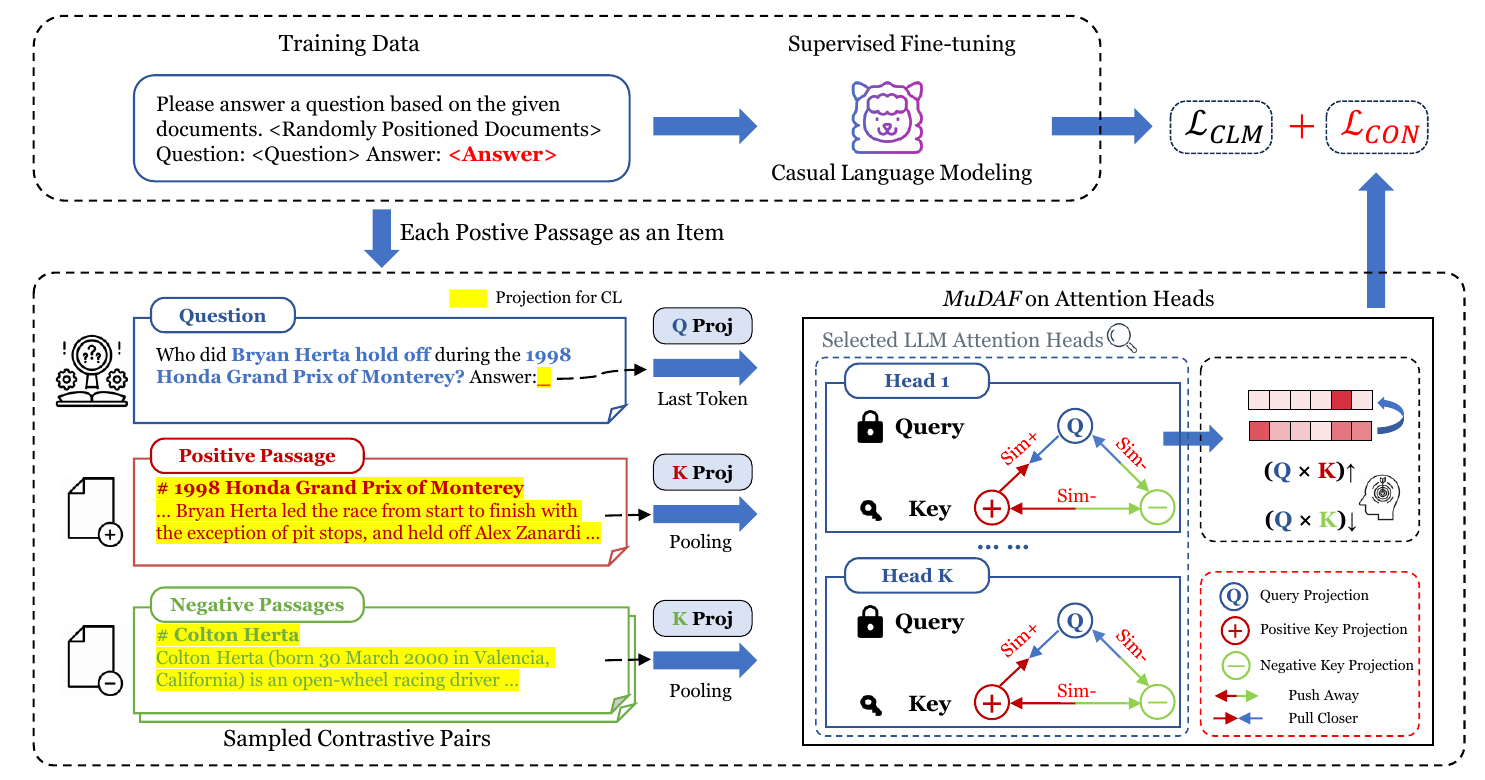}
    % \caption{\FMT~method is able to align the multilingual representations of truthful statements in semantic space, making LLMs more truthful to multilingual answer questions.}
    \caption{An overview of our proposed method. The goal of~\method~is to adjust the similarity between the \texttt{Query} features from the question and the \texttt{Key} features from the passages, thus making attention heads allocate more attention weights in relevant information and reducing distractions. CL means contrastive learning.}
    % , thus providing more accurate response.
    \label{fig:overview}
\end{figure*}

%%%%%%%%%%%%%%%%%% need to be major revised %%%%%%%%%%%%%

\paragraph{Contrastive Learning on Generative Models.} As a self-supervised training technique, contrastive learning~\citep{chopra2005learning, hadsell2006dimensionality, robinson2021contrastive} has been widely leveraged in NLP tasks such as sentence embedding~\citep{gao2021simcse,li2023multi,zhang2022fine}. With the advancement of generative language models~\citep{radford2019language}, contrastive learning has also exhibited great potential in decoder-only architectures to achieve better hidden expressiveness~\citep{su2022a, jain-etal-2023-contraclm, yan2024contrastive}. For long-context tasks,~\citet{caciularu-etal-2022-long} utilize contrastive learning to explicitly discriminate representations of supporting evidence sentences from negative ones in long-context QA.~\citet{wu2024reducing} also leverage contrastive learning to align representations of different contexts. However, our method applies contrastive learning inside the attention head components instead of sequence representations. To the best of our knowledge, we are the first to show the effectiveness of optimizing attention distributions by adjusting the similarity between query and key projections at the head level directly.

\section{Method}

In this section, we introduce our proposed method. We start by investigating the relationship between the performance of an LLM in MDQA and the ability of its attention heads for information retrieval to identify its retrieval heads~(\textsection \ref{subsec:IR}). We then discuss the details of our method~(\textsection \ref{subsec:CL}). An overview of our approach is provided in Figure~\ref{fig:overview}.

\subsection{Attention Heads Responsible for IR}
\label{subsec:IR}

Information retrieval~(IR) here means recognizing required information from noisy input context.~\citet{wu2024retrieval} have proven the existence of retrieval heads in the NIAH test~(i.e., NIAH retrieval heads), which implement the conditional copy algorithm and redirect information from the input to the output. However, it remains unclear whether these retrieval heads function similarly in other long-context tasks, such as MDQA, where LLMs are required to retrieve relevant information from previous passages to answer a given question rather than simply repeating patterns found in the context.

\begin{figure}[thp]
    \centering
    \includegraphics[width= 0.48\textwidth]{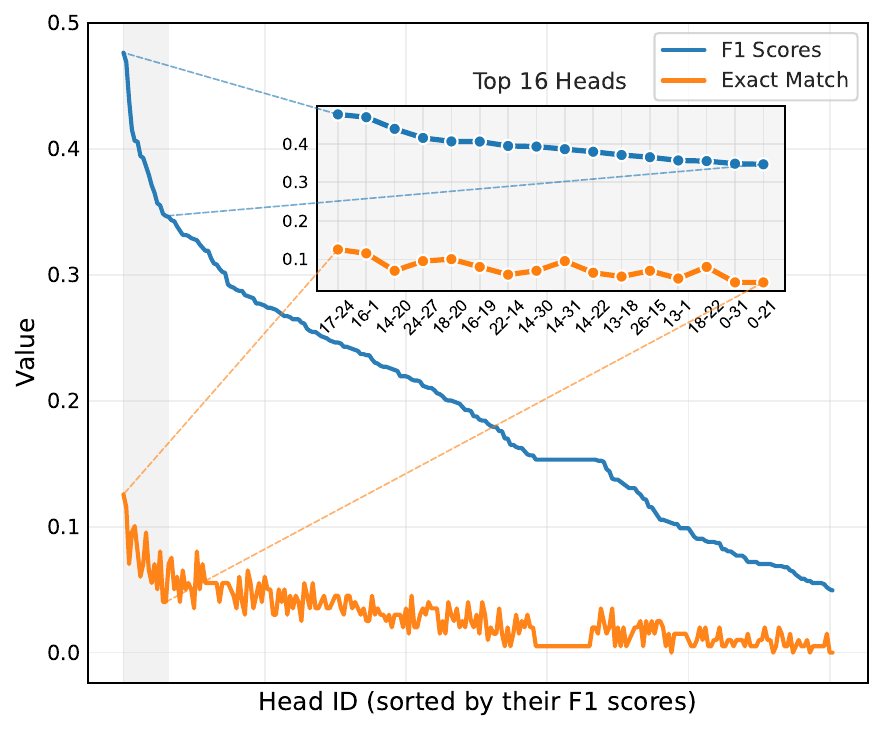}
    \caption{The F1 and EM retrieval scores for attention heads of Llama3.1-8B. We list top 16 retrieval heads ranked by their F1 scores in the inner graph.}
    \label{fig:fig_em_f1}
\end{figure}

First, we manually labeled golden passages for all questions in the HotpotQA subset of LongBench. Our annotation pipeline can be found in Appendix~\ref{app:label}. We define the retrieval score of a given attention head as their ability to attend to golden passages among all input passages. Formally, for each question $q$, we have several relevant golden passages $P_G$ and quite a few irrelevant passages $P_I$. We mixed all $P_G$ and $P_I$ into a long input context $\mathcal{C}$ in random order, then we concatenated the question $q$ to the end of $\mathcal{C}$ to obtain the final prompt $\mathcal{P}$. For each attention head, we calculated its attention score over all input passages by summing the attention scores between the last token of the input and all tokens in the corresponding passages. The passage whose attention score was higher than a given threshold $\epsilon$ would be considered an attended passage $P_{A_h}$ of attention head $h$. We then calculated the F1 score and EM score based on $P_G$, $P_I$ and $P_A$. Figure~\ref{fig:fig_em_f1} presents the curves of F1 scores and EM scores of all attention heads in Llama-3.1-8B~\cite{llama3_1}, ranked in descending order of F1 scores. The final retrieval score $\mathcal{R}_h$~($0 \leq \mathcal{R}_h \leq 1$) of an attention head $h$ is the average F1 score on all HotpotQA test cases. The formula for calculating $\mathcal{R}_h$ is as follows:
\begin{equation}
    \mathcal{R}_h = mean(F1\_Score(P_G,P_I,P_A))
\end{equation}
Details about the calculation process can be found in Appendix~\ref{app:retrieval_score}.

\begin{figure}[thp]
    \centering
    \includegraphics[width= 0.45\textwidth]{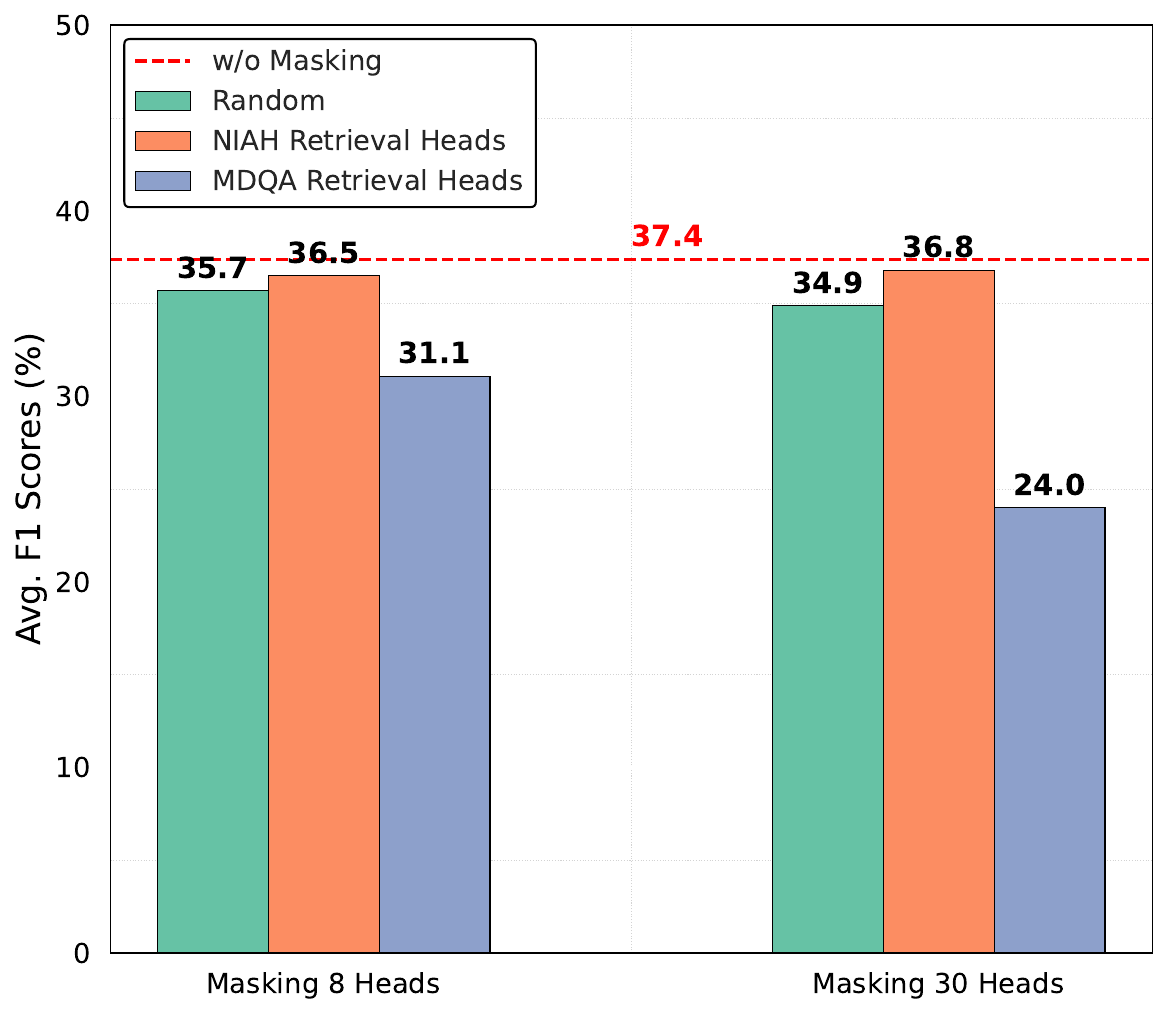}
    % \caption{\FMT~method is able to align the multilingual representations of truthful statements in semantic space, making LLMs more truthful to multilingual answer questions.}
    \caption{Average performance of Llama3.1-8B on LongBench with different masking strategies. In this experiment, we used the MDQA subset of LongBench, including HotpotQA, 2WikiMQA and MuSiQue. Masked retrieval heads were also randomly selected from the set of retrieval heads, and the final results were obtained by averaging three independent experimental runs.}
    % , thus providing more accurate response.
    \label{fig:mask_exp}
\end{figure}

We found that retrieval scores decline smoothly from the strongest heads to the weakest heads, making it difficult to distinctly classify them as either "strong" or "weak" based on a clear threshold. For convenience, \textbf{we consider the top 50 attention heads~(about 5\% of the total) as retrieval heads}. To confirm the essentiality of these retrieval heads, we then carried out further masking experiments. As shown in Figure ~\ref{fig:mask_exp}, the performance on LongBench is severely damaged when strong MDQA retrieval heads are masked, showing that they play a vital role in multi-document modeling. In addition, the model's performance exhibits smaller fluctuations when random attention heads or NIAH retrieval heads are masked. This experimental result gives us a reliable direction for selecting which attention heads to optimize. In other words, we may assume that only by optimizing attention heads with a strong enough retrieval capability can we improve the long-context modeling ability of the model.
% More detailed analysis of MDQA retrieval heads can be found in Appendix~\ref{app:retrieval_score}. 
% 放一个例子

% 用来说明的是：attention 的分布 => NIAH and MDQA
% 用两个 head 对比？可是这样的比较对于后续的有意义吗？
% 后续要说明的是某一层的 retrieval score 提高了，能够改善对于相关信息的 retrieval 能力，而 QA base 的就没有
% 说明区别？
% theory of our proposed method:
% To enhance
% Attention 与 QA 正确性的证明
% perform copy-paste in NIAH test, we hold the 
% mainly did their experiments on NIAH test, but we 
% 

% \subsection{Attend to What You Need}

\subsection{Contrastive Learning for Optimizing Attention Heads}
\label{subsec:CL}

We have proven that MDQA retrieval heads can offer reliable key information in input context, helping LLMs leverage golden information to answer the given question. Therefore, we propose~\method, a method based on joint training of casual language modeling~(CLM) and contrastive learning, aiming at enhancing MDQA retrieval heads for a better focus on relevant context and reducing the distraction caused by irrelevant content. 

\paragraph{Preliminary.} To better describe our method, we first define some universal notations and variables:
\begin{itemize}
    \item $\mathcal{H}$: attention head set of the model. $\mathcal{H}_i$ means the attention head set in the $i$th layer of the model.
    \item $N$: the number of layers that the model has.
    \item $k$ and $\mathcal{K}$: $k$ indicates the index of a certain passage. $\mathcal{K}$ means the number of passages for a given example.
    \item $[h]$: a superscript that denotes a specific attention head.
\end{itemize}
\paragraph{Attention Simplification for MDQA.} We perform a simplification to the attention mechanism in the MDQA setting that makes it easy to understand our optimization goal. 

Let $\mathcal{C}$ represent the holistic long context, which consists of golden passages $P_G$, irrelevant passages $P_I$ and the question $q$. We assume that LLMs can better answer a question if they attain more hidden information directly from golden passages $P_G$. To describe this information aggregation process more concisely, we define the hidden information of passage $P_k$ stored at layer $i$ as $\mathcal{I}_{i,k}$, and we only consider the attention distribution of the last input token. We simply define a substitutive attention weight of head $h$ for a passage $P_k$ as $\mathcal{A}^{[h]}_k$. Meanwhile, for other tokens in the input, we define their attention weight and hidden information as $\mathcal{A}^{[h]}_{\circ}$ and $\mathcal{I}_{i-1}$. Then we can simply denote the information obtained by layer $i$ as follows:
\begin{equation}
    \begin{aligned}
        \mathcal{I}_i &= \mathcal{O}_i\bigg( \text{concat}_h^{\mathcal{H}_i} \bigg\{ & \sum_k^\mathcal{K} \mathcal{A}^{[h]}_k \cdot \mathcal{I}_{i-1,k} \\
        &\quad & + \mathcal{A}^{[h]}_{\circ} \mathcal{I}_{i-1} \bigg\} \bigg)
    \end{aligned}
\end{equation}
where $\mathcal{O}_i$ represents the output projection module of layer $i$, $\mathcal{H}_i$ represents the attention heads set of layer $i$, $\mathcal{A}^{[h]}_k$ is the attention score on passage $P_k$, while $\mathcal{A}^{[h]}_\circ$ represents the sum of attention scores of all other non-passage tokens.
To make LLMs gain more information from golden passages and mitigate the disturbance by irrelevant passages, our optimization goal is to increase the attention weights assigned to golden passages \(P_G\) while reducing the attention weights assigned to irrelevant passages \(P_I\). Formally, a multi-head attention score~\citep{vaswani2017attention} between the last input token $t$ and a token $i$ inside a passage can be expressed as:
\begin{equation}
    \begin{aligned}
    \text{attn}^{[h]}_i &= \bigg[\text{softmax} \bigg( \dfrac{Q^{[h]}_t(K^{[h]})^T}{\sqrt{d}} \bigg)\bigg]_i \\
                      &= \frac{\exp \big( \dfrac{Q^{[h]}_t (K^{[h]}_i)^T}{\sqrt{d}} \big)}{\sum_{j} \exp \big( \dfrac{Q^{[h]}_t (K^{[h]}_{j})^T}{\sqrt{d}} \big)}.
    \end{aligned}
\end{equation}
Intuitively, the attention score can be raised through aligning the representations of $Q^{[h]}_t$ and $K^{[h]}_i$, as well as pushing apart $Q^{[h]}_t$ and $K$ projections of other tokens in the embedding space.
Therefore, we can leverage contrastive learning~\citep{hadsell2006dimensionality} at the head level inside the attention mechanism to adjust the attention allocation. In our approach, we argue that aggregating the representations of all tokens in a passage for contrastive learning is as effective as performing contrastive learning on the representation of each individual token. Hence, we perform an average pooling operation on $K$ projections of all tokens to obtain the overall $K$ representation of a passage. Then the attention weight $\mathcal{A}^{[h]}_k$ can be written as:
\begin{equation}
    \mathcal{A}^{[h]}_k = \sum_{t' \in P_k} \text{attn}^{[h]}_{t'} \approx \text{attn}^{[h]}_{P_k} \\
\end{equation}
\begin{equation}
\text{attn}^{[h]}_{P_k} = \text{softmax}\bigg(\frac{Q^{[h]}_t \left( \frac{1}{|P_k|} \sum_{t' \in P_k} K^{[h]}_{t'} \right)^T}{\sqrt{d}}\bigg)
\end{equation}
We denote the pooled $K$ representation of passage $P_k$ as $K_{P}$:
\begin{equation}
    K^{[h]}_P = \frac{1}{|P_k|} \sum_{t' \in P_k} K^{[h]}_{t'}
\end{equation}
\paragraph{Objective of Contrastive Learning.} To magnify the attention weight allocated to golden passages, the objective of the contrastive learning is to maximize the similarity between $Q^{[h]}_t$ and $K^{[h]}_{P_G}$, while pushing apart the representations of $Q^{[h]}_t$ and $K^{[h]}_{P_I}$. Therefore, the loss function can be presented as follows:
\begin{equation}
    \mathcal{L}_{\text{CON}} = - \sum_{h} \log \frac{e^{(\text{sim}(Q^{[h]}_t, K^{[h]}_{P_k})/\tau)}}{\sum_{P_j \in P} e^{(\text{sim}(Q_t^{[h]}, K^{[h]}_{P_j})/\tau)}}.
\end{equation}
where $P=P_G\cup P_I$, sim($\cdot$,$\cdot$) denotes the cosine similarity function and $\tau$ is a temperature hyperparameter. Finally, we combine $L_{\text{CON}}$ with a Causal Language Modeling~(CLM) loss function as the overall loss function:
\begin{equation}
    \mathcal{L} = \mathcal{L_{\text{CLM}}} + \lambda \mathcal{L}_{\text{CON}}
\end{equation}
where $\lambda$ is a hyperparameter to control the weight of $\mathcal{L}_{\text{CON}}$.
% 需要一个方法的图

% 主讲 QK-level 的对比学习方法，去激活 retrieval heads.
% 首先简单讲讲 attention 的基本概念，然后引出在 Q-K Level 上进行对比学习的可行性。
% 为了强化 passage-level retrieval head，我们希望这些 head 能够 retrieve 到正确的信息，也就是 attend 到正确的和 question 相关的 passages。因此，我们提出了 Long-PRX，如图所示， Long-PRX 通过对比学习优化 retrieval heads 的 Q_projection 和 K_projection ，同时也不冻结模型的其他权重，并辅以正常的 QA_loss，优化了模型的 attention 分布的同时，也保证模型不丧失基本的模型能力。

% 可能更多attention head 参与训练时效果不显著是由于没有加入 CoT，部分 reasoning head 的能力没有跟上

\section{Experiments}

\subsection{Setup}

\paragraph{Benchmarks} We evaluated our fine-tuned models on LCQA datasets, including both multi-document question answering and single-document question answering subsets from LongBench~\citep{bai-etal-2024-longbench} and ZeroSCROLLS benchmarks~\citep{shaham-etal-2023-zeroscrolls}. LongBench is a bilingual and multitask benchmark for long-context understanding. Among its subsets, we included HotpotQA~\citep{yang2018hotpotqa}, 2WikiMQA~\citep{xanh2020_2wikimultihop}, MuSiQue~\cite{10.1162/tacl_a_00475} and Qasper~\citep{dasigi2021dataset}. ZeroSCROLLS also provides various datasets for evaluating models' capabilities in synthesizing information over long texts, and we included the MuSiQue and Qasper subsets from it. The statistics of benchmarks are listed in Table~\ref{tab:benchmarks}.

\begin{table}[tp]
\resizebox{0.49 \textwidth}{!}{
\begin{tabular}{lcrr}
\toprule
\textbf{Dataset}            & \textbf{Type}          & \textbf{Avg \#Words} & \textbf{\#Items} \\
\midrule
\textit{LongBench} &               &        &        \\
HotpotQA           & Multi-Doc QA  & 9,151                          & 200                        \\
2WikiMQA           & Multi-Doc QA  & 4,887                          & 200                        \\
MuSiQue            & Multi-Doc QA  & 11,214                         & 200                        \\
Qasper             & Single-Doc QA & 3,619                          & 200                        \\
\midrule
\textit{ZeroSCROLLS}        &               &                                &                            \\
MuSiQue            & Multi-Doc QA  & 1,749                          & 500                        \\
Qasper             & Single-Doc QA & 3,531                          & 500  \\                     
\bottomrule
\end{tabular}
}
\caption{An overview of the benchmark statistics. The metric for these datasets are all F1 scores. Note that the MuSiQue subset in ZeroSCROLLS contains unanswerable questions and models should refused to answer them.}
\label{tab:benchmarks}
\end{table}

% Statistics and more information about tested datasets can be found in Appendix~\ref{app:datasets}.

% \begin{table}[hp]
% \resizebox{0.49 \textwidth}{!}{
% \begin{tabular}{lccrr}
% \toprule
% \textbf{Dataset}            & \textbf{Type}          & \textbf{Metric} & \textbf{Avg \#Words} & \textbf{\#Items} \\
% \midrule
% \textit{LongBench} &               &        &            &        \\
% HotpotQA           & Multi-Doc QA  & F1                         & 9,151                          & 200                        \\
% 2WikiMQA           & Multi-Doc QA  & F1                         & 4,887                          & 200                        \\
% MuSiQue            & Multi-Doc QA  & F1                         & 11,214                         & 200                        \\
% Qasper             & Single-Doc QA & F1                         & 3,619                          & 200                        \\
% \midrule
% \textit{ZeroSCROLLS}        &               &                            &                                &                            \\
% MuSiQue            & Multi-Doc QA  & F1                         & 1,749                          & 500                        \\
% Qasper             & Single-Doc QA & F1                         & 3,531                          & 500  \\                     
% \bottomrule
% \end{tabular}
% }
% \caption{An overview of the benchmark statistics. Note that the MuSiQue subset in ZeroSCROLLS contains unanswerable questions and models should refused to answer them.}
% \label{tab:benchmarks}
% \end{table}

\begin{table*}[htbp]
\centering
\resizebox{0.98 \textwidth}{!}{
\begin{tabular}{l|cccccccl}
\toprule
\multirow{2}{*}{\textbf{Models}} & \multicolumn{4}{c}{\textbf{LongBench}}    &    & \multicolumn{2}{c}{\textbf{ZeroSCROLLS}} & \multirow{2}{*}{\textbf{Avg.}} \\ \cmidrule{2-5} \cmidrule{7-8}
                                 & HotpotQA & 2WikiMultihopQA & MuSiQue & Qasper & & MuSiQue             & Qasper             &                                \\ \midrule
GPT-4o~\citep{gpt4o}                   &   68.3   &   49.5     &  39.8  & \underline{46.1}   & & \underline{59.5}                & \underline{48.7}               & \underline{52.0}                           \\ 
GPT-3.5-Turbo~\cite{chatgpt}           &   51.6   &   37.7     &  26.9  & 43.3   & & 52.0               & 27.1               & 39.8                           \\ \midrule
% Llama-3.1-8B                     & 48.9     & 39.8            & 32.6    & 30.8   & & 20.3                & 35.6               & 34.6                           \\ 
%Llama3-ChatQA-2-8B                 & 52.5     & 41.8            & 38.9    & 28.5   & &                 &                &                            \\ 
FILM-7B~\cite{an2024make}              &   62.1   &   47.0     &  39.0   & 42.2   & &       \textbf{35.2}        &     \textbf{54.7}    &     46.7     \\ 
ChatQA-2-8B~\citep{xu2024chatqa}       &   52.5   &   41.8     &  38.9   & 28.5   & &       27.3         &     47.9       &                         39.5 \\ 

ProLong-8B-64k~\cite{gao2024train}     &   43.0   &   24.9     &  21.3   & 27.8   & &       25.7         &     36.7     &                           29.9  \\ 
Llama3.1-8B-Instruct~\cite{llama3_1}   &   54.7   &   44.0     &  32.8   &  \textbf{44.7}  & &   29.1    &   51.8      &                  42.8  \\ 
\rowcolor{pink!50} Llama3.1-8B-\textit{Vanilla-SFT}       &   46.8   &   50.5     &  28.9   & 29.2  & & 30.1                & 41.6               & 37.8                           \\ 
Llama3.1-8B-\textit{MuDAF-weak}        & 62.5     & 53.8            & 43.1    & 34.9   & & 23.2                & 41.8               &              43.2                 \\
% \rowcolor{pink!50} Llama3.1-8B-\method~\textbf{(ours)}    & \textbf{69.6}~\textbf{(+22.8\%)} & \textbf{66.2}~\textbf{(+15.7\%)} & \textbf{48.2}~\textbf{(+19.3\%)}  & 40.0~\textbf{(+10.8\%)}  & & 31.2~\textbf{(+1.1\%)}                & 47.9~\textbf{(+6.3\%)}               & \textbf{50.5}~\textbf{(+12.7\%)}                           \\
\rowcolor{pink!50} Llama3.1-8B-\method~\textbf{(ours)}    & \textbf{69.6} & \textbf{66.2} & \textbf{48.2}  & 40.0  & & 31.2                & 47.9               & \textbf{50.5}                           \\
\bottomrule
\end{tabular}

% \href{https://huggingface.co/princeton-nlp/Llama-3-8B-ProLong-64k-Base}{https://huggingface.co/princeton-nlp/Llama-3-8B-ProLong-64k-Base}
%                   &      &            &     &    & &                 &              &                            \\ 

% Llama3.1-8B-\textit{Vanilla-SFT}       &   62.1   &   55.2     &  40.3   & 36.7  & & 30.1                & 41.6               & 44.3                           \\ % better answer extraction
}
\caption{F1 scores~(\%) on all tested datasets. \underline{Underlined} numbers denotes the best performance among all listed models. \textbf{Bold} numbers indicates the best performance of tested open source models and our models. \textit{Vanilla-SFT} means training the foundation model without $\mathcal{L}_\text{CON}$. \textit{MuDAF-weak} means that we apply our method to weak attention heads (\textsection \ref{sec:analysis}). \method~achieves better performance among all tested datasets compared with the~\textit{Vanilla-SFT} baseline.}
\label{tab:main_results}
\end{table*}

 % Vicuna-v1.5-7b-16k~\citep{NEURIPS2023_91f18a12},
 % FILM-7B~\citep{an2024make}, 
\paragraph{Baselines and Foundation Models} We compared our approach to several popular and strong long-context LLMs, including GPT-3.5 Turbo~\citep{chatgpt}, GPT-4o~\citep{gpt4o}, FILM-7B~\citep{an2024make}, ChatQA-2-8B~\citep{xu2024chatqa}, ProLong-8B-64k\footnote{\href{https://huggingface.co/princeton-nlp/Llama-3-8B-ProLong-64k-Base}{https://huggingface.co/princeton-nlp/Llama-3-8B-ProLong-64k-Base}}~\citep{gao2024train}, and Llama3.1-8B-Instruct-128k~\cite{llama3_1}. In this paper, our method are applied to the foundation model Llama-3.1-8B~\cite{dubey2024llama}, which has 128K context length. We also include an intuitive vanilla supervised fine-tuning~(\textit{Vanilla-SFT}) baseline that fine-tuning the foundation model only with $\mathcal{L}_{\text{CLM}}$ loss.

\paragraph{Training Data} We adopted the training dataset of HotpotQA~\citep{yang2018hotpotqa} with additional hard negative passages. Here negative samples are enriched by similar passages collected from the work by~\citet{jiang2024longrag}, where they grouped multiple
Wikipedia documents through hyperlinks. Thus, we expanded negative passages utilizing the group of existing passages in the original set. 

% The average input context length of training data is 8K.

\subsection{Implementation Details}

\paragraph{Target Heads Selection} To implement \method, we first select several strong MDQA retrieval heads for contrastive learning. Since we have got retrieval scores of all attention heads in section \textsection \ref{subsec:IR}, we could simply select retrieval heads with the highest retrieval score. However, we found that this greedy strategy is not the best and not robust. We thus use a weighted random selection algorithm that randomly picks attention heads based on their retrieval scores. More specifically, given attention heads set $\mathcal{H}$ and their retrieval scores $\{\mathcal{R}_h\}$, the probability $P(h)$ of selecting the attention head $h$ is computed as:
$$
P(h) = \dfrac{e^{\mathcal{R}_{h}/\tau}}{\sum_{h'\in \mathcal{H}} e^{R_{h'}/\tau}}
$$
where $\tau>0$ is a temperature parameter (e.g., 0.05), $\mathcal{H}$ denotes all attention heads. We set the number of selected target heads to 8.
% 方法受制于分析处得到的结论

\paragraph{Fine-tuning Details} One training sample of contrastive learning contains one golden passage to the given question and many negative passages. We separately compute $Q$ projection for the last token of the question and pooled $K_P$ projection for all passages. During the similarity calculation, we concatenate corresponding representations from all selected attention heads and calculate the overall cosine similarity between them. We found that this implementation is more stable and also effective compared with calculating the similarity for each attention head separately. For \textit{Vanilla SFT}, the order of input passages is randomly shuffled before forming an MDQA input and computing CLM loss. More details can be found in Appendix~\ref{app:imple}.

\subsection{Main Results}

% 结果上: 1. 点数上涨
%        2. retrieval 的 F1 分数上升。并有对比。
%           维度上有：各类方法的对比（平均F1, 对应头的 F1, 对应层的 F1）
%        3. 

% \paragraph{\method~significantly enhance model's LCQA performance.}
\paragraph{Enhancement on LCQA Performance.} Our method significantly enhances the model's LCQA performance. Table~\ref{tab:main_results} compares the performance of our method with other baselines. \textit{MuDAF} shows great potential in enhancing the LCQA performance of models, getting \textbf{+12.7\%} improvement on average scores compared with the \textit{Vanilla-SFT} baseline. Meanwhile, our method is also effective on single-document question-answering datasets~(e.g., Qasper), indicating that our method is also robust in enhancing the retrieval capabilities of LLMs in one long document. Moreover, our method achieves comparable performance to that of GPT-4o, and even performs better on some datasets, proving the effectiveness of our method. Note that one in five questions of the MuSiQue subset from ZeroSCROLLS are unanswerable, which may have affected the performance of our method on this dataset.

\begin{figure*}[thp]
    \centering
    \includegraphics[width= 0.90\textwidth]{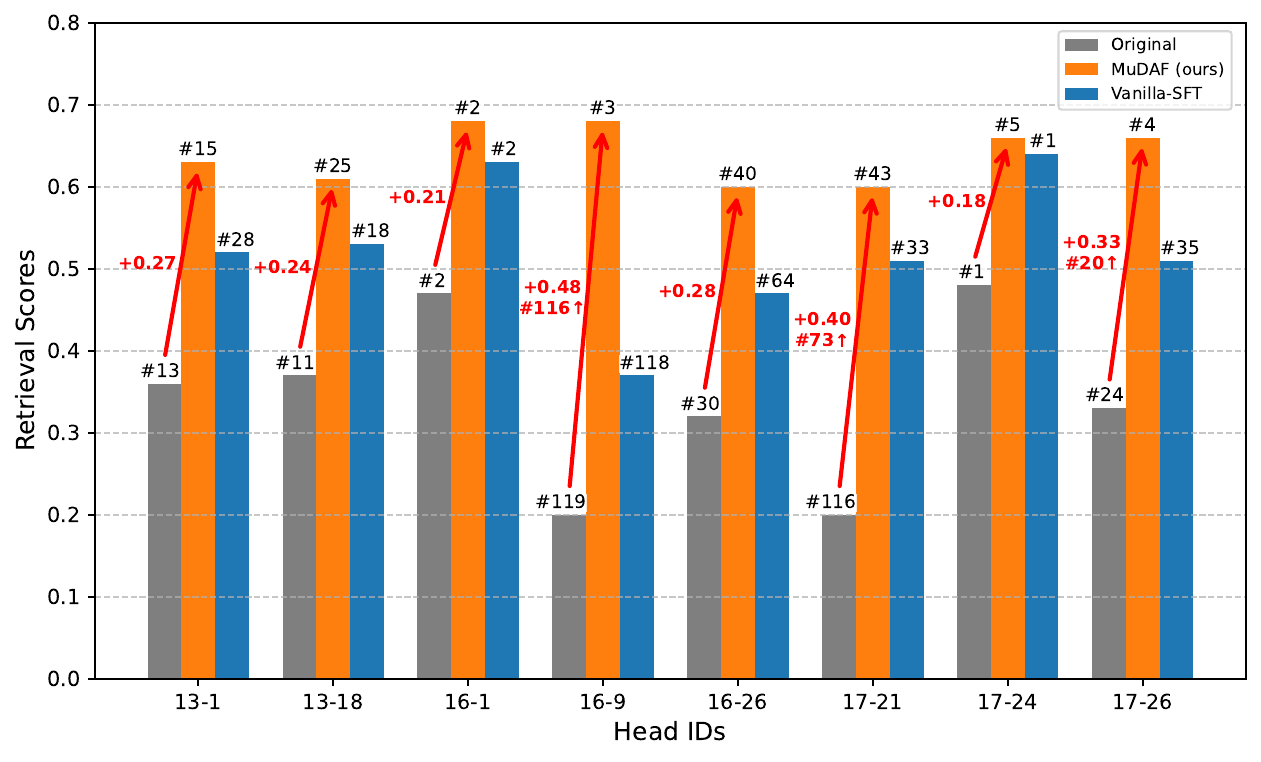}
    \caption{Comparison of enhanced retrieval capabilities in selected attention heads. We annotate the rank of each attention head among all heads above the bar~(i.e., \#x).}
    % , thus providing more accurate response.
    \label{fig:head_improve}
\end{figure*}

\paragraph{Achieving More Focused Retrieval Heads.}
% 需要 attention 的样例分析。
% 需要 
% \paragraph{Trained heads are obviously more focused on golden content.}
% 展示出 头的 优化，需要一个表
Besides improvements on QA performance, we want to examine whether~\method~can genuinely strengthen the retrieval capabilities of target attention heads. Therefore, we make a comparison of target attention heads' retrieval scores between the original model,~\method~and \textit{Vanilla-SFT}. As shown in Figure~\ref{fig:head_improve}, all target attention heads get significant enhancement from~\method~on retrieval scores compared with the original model, and the optimization gains are also substantially larger than those observed with the \textit{Vanilla-SFT} baseline. For example, head 16-9 achieves a \textbf{+0.48} improvement in retrieval scores through~\method, elevating its ranking from \textbf{the 119th to the 3rd} place instantaneously. In contrast, the \textit{Vanilla-SFT} baseline brings few improvements to its ranking. It is worth noting that stronger heads often achieve smaller improvements, indicating that it is easier to enhance those attention heads in the middle part.

% Add one more paragraph? [TODO]

\subsection{Analysis}
\label{sec:analysis}
% We did further ablation studies to investigate the effectiveness of selecting strong MDQA retrieval heads and the influence of the number of selected attention heads.

% \begin{figure}[thp]
%     \centering
%     \includegraphics[width= 0.48\textwidth]{head_improve.pdf}
%     \caption{Comparison of enhanced retrieval capabilities in selected attention heads. }
%     % , thus providing more accurate response.
%     \label{fig:head_improve}
% \end{figure}

\begin{figure}[thp]
    \centering
    \includegraphics[width= 0.46\textwidth]{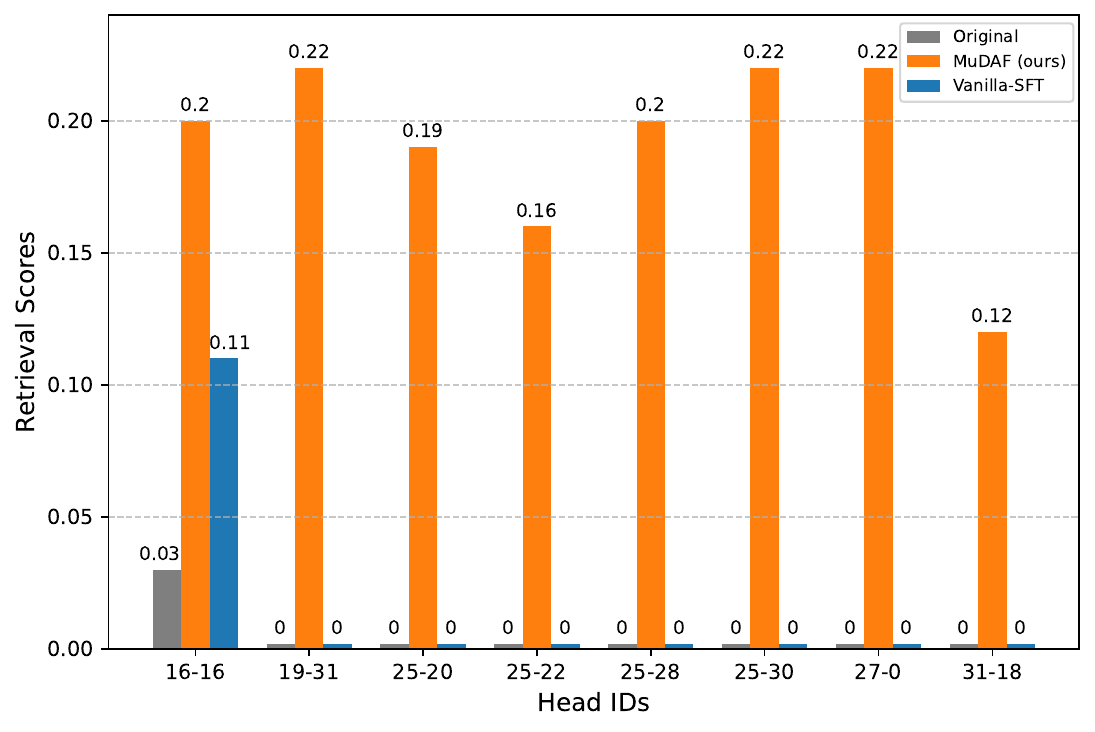}
    \caption{Enhanced retrieval capabilities when applying \textit{MuDAF} to weak attention heads.}
    % , thus providing more accurate response.
    \label{fig:weak_head_improve}
\end{figure}

\paragraph{Effectiveness on Weak Heads.} In this paper, we regard attention heads with low retrieval scores~(i.e., $\mathcal{R}_h$ < 0.1) as weak attention heads. Due to the promising improvements of both performance and retrieval scores when applying~\method~to strong retrieval heads, we are curious about whether our method could transform weak attention heads into heads with a certain retrieval capability. Therefore, we randomly selected attention heads whose retrieval scores are nearly zero for optimization. As shown in Table~\ref{tab:main_results}, the overall performance is relatively weak, but it is still an improvement compared with the \textit{Vanilla-SFT} baseline. We further calculated their retrieval scores after the training stage. As illustrated in Figure~\ref{fig:weak_head_improve},~\method~can exactly enhance the retrieval capabilities of these weak heads, while \textit{Vanilla-SFT} does nothing in it. This phenomenon manifests that we could adjust the attention pattern of one head through~\method~towards retrieval heads even though they are extremely weak attention heads in the original model. But at the same time, obviously we can hardly achieve the same performance as strong retrieval heads do since their attention values are still relatively low in the middle context.

% and how such improvements on weak heads help the model recognize relevant passages remains to be explored 
% (需要这几句吗?)
% We calculated the average improvements

% narrative 不够好

\paragraph{Whole-layer Optimization.}
% 体现对比，依然用表现和F1 score
% 需要吗？
Although we are focusing on target attention heads in previous experiments, the model parameters of other parts are not frozen, making them possible to get improved as well. So we also calculated the attention scores after the training for other attention heads that are not directly selected for the contrastive learning. Surprisingly, we found that most attention heads within the same layer can also be optimized when incorporating at least one attention head in the contrastive learning process. We discuss more about this interesting phenomenon in Appendix~\ref{app:whole_layer}.

% This may be caused by the Group Query Attention~(GQA)~\citep{}

\begin{figure}[hp]
    \centering
    \includegraphics[height=0.35\textwidth]{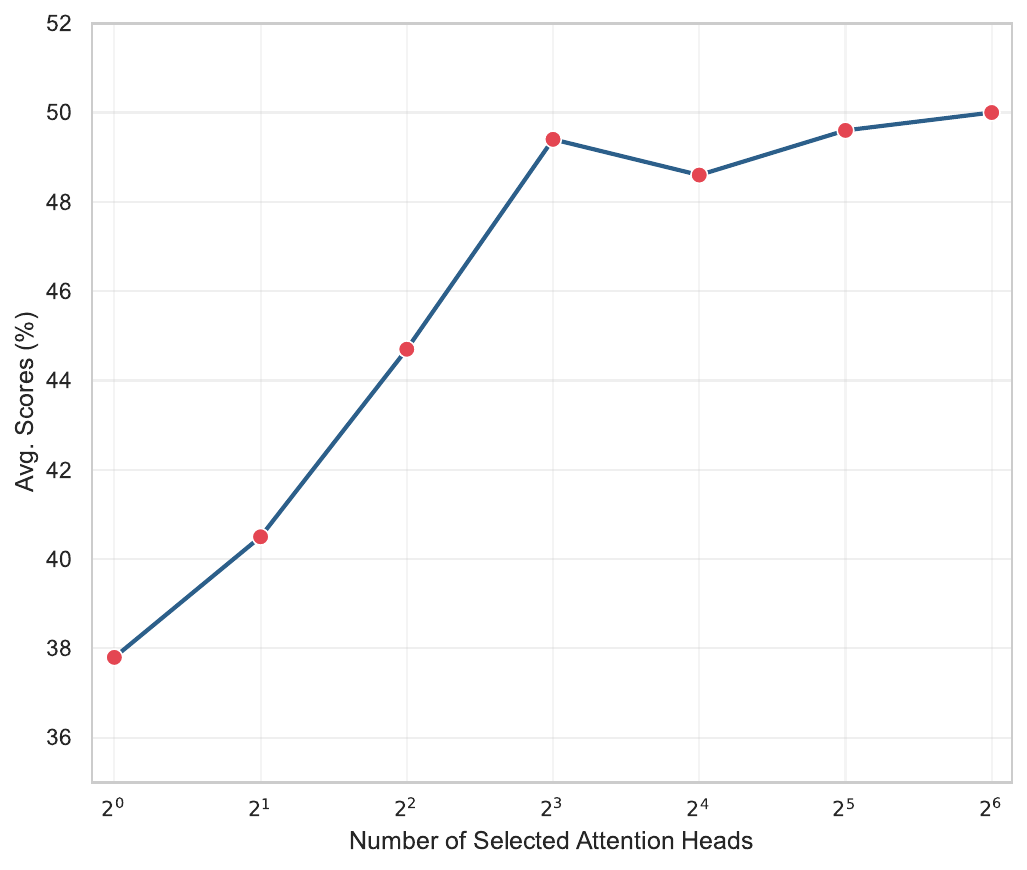}
    \caption{Ablation on the number of selected attention heads. In this ablation, we select attention heads according to the retrieval score from high to low.}
    % , thus providing more accurate response.
    \label{fig:num_ablation}
\end{figure}

% \paragraph{A Small Number of Selected Heads is Enough.} 
\paragraph{Bottleneck When Scaling the Number of Target Heads.}
We wonder if it is possible to get a stronger model through applying \method~to more attention heads. Unfortunately, we discovered a bottleneck when scaling the number of trained attention heads. As depicted in Figure~\ref{fig:num_ablation}, we did not see much improvement if we consistently increased the number of selected attention heads beyond 8 heads. Furthermore, if we engage all attention heads in contrastive learning, the training will become unstable and struggle to converge, leading to a collapse of the overall performance. We remain the investigation of this bottleneck as future research.
% 一个图即可

\paragraph{Case study.} We present a case study to show the effectiveness of~\method~to make the model be more focused on relevant passages within a long context. Through our observations, our approach can correct incorrect attention assignments while enhancing attention weights allocated to golden documents, exhibiting its potential to convert weak attention heads into retrieval-capable attention heads. Details are provided in Appendix~\ref{app:case_study}.

\section{Conclusion}

% In this paper, we aim at optimizing certain attention heads and helping them be more focused on relevant content in LCQA tasks. We find special retrieval heads in the MDQA setting that are different from those found in the NIAH test. To enhance these retrieval heads, we propose~\method, a method that can significantly improve the retrieval capabilities of certain attention heads regardless of whether they are strong or weak heads. Consequently, LLMs' performance in LCQA tasks, especially MDQA, are also getting enhanced. Our method and experiments draw a promising roadmap and provide valuable insights in utilizing contrastive learning to optimize attention distribution at the head level in MDQA tasks.

In this paper, we focus on optimizing specific attention heads to enhance their ability to concentrate on relevant content in LCQA tasks. Our analysis reveals the existence of specialized retrieval heads in the MDQA setting that differ from those found in the NIAH test. To improve these retrieval heads, we introduce~\method, an approach that significantly enhances the retrieval capabilities of attention heads in MDQA regardless of their initial strength. Consequently, the performance of LLMs in LCQA tasks gets remarkable improvements as well. Our method and experiments draw a promising roadmap and provide valuable insights in utilizing contrastive learning to optimize the attention distribution at the head level in MDQA tasks.

\section*{Limitations}
% 可以做 early exist
% 没有详细分析推理过程中模型从单独一篇文章中获取的信息变化。
% 可以考虑更多的 head

Although we see improvements on attention scores and the LCQA performance through our method, it is still hard to explain the relationship between optimizing a certain head's attention distribution and the final output of the model, since other attention heads also engage in reasoning and making the final response. 

Moreover, our approach can be affected by the position of the question, which means the model can better retrieve relevant documents in the input through attention focusing only when the question is at the end of the input sequence. It is of great importance if we can design a more robust method to mitigate such positional bias.

% \section*{Ethical Considerations}

\bibliography{custom}

\appendix

\section{Retrieval Head Detection}
\subsection{LongBench Annotation}
\label{app:label}

\begin{figure*}[hp]
    \centering
    \includegraphics[width= 0.88\textwidth]{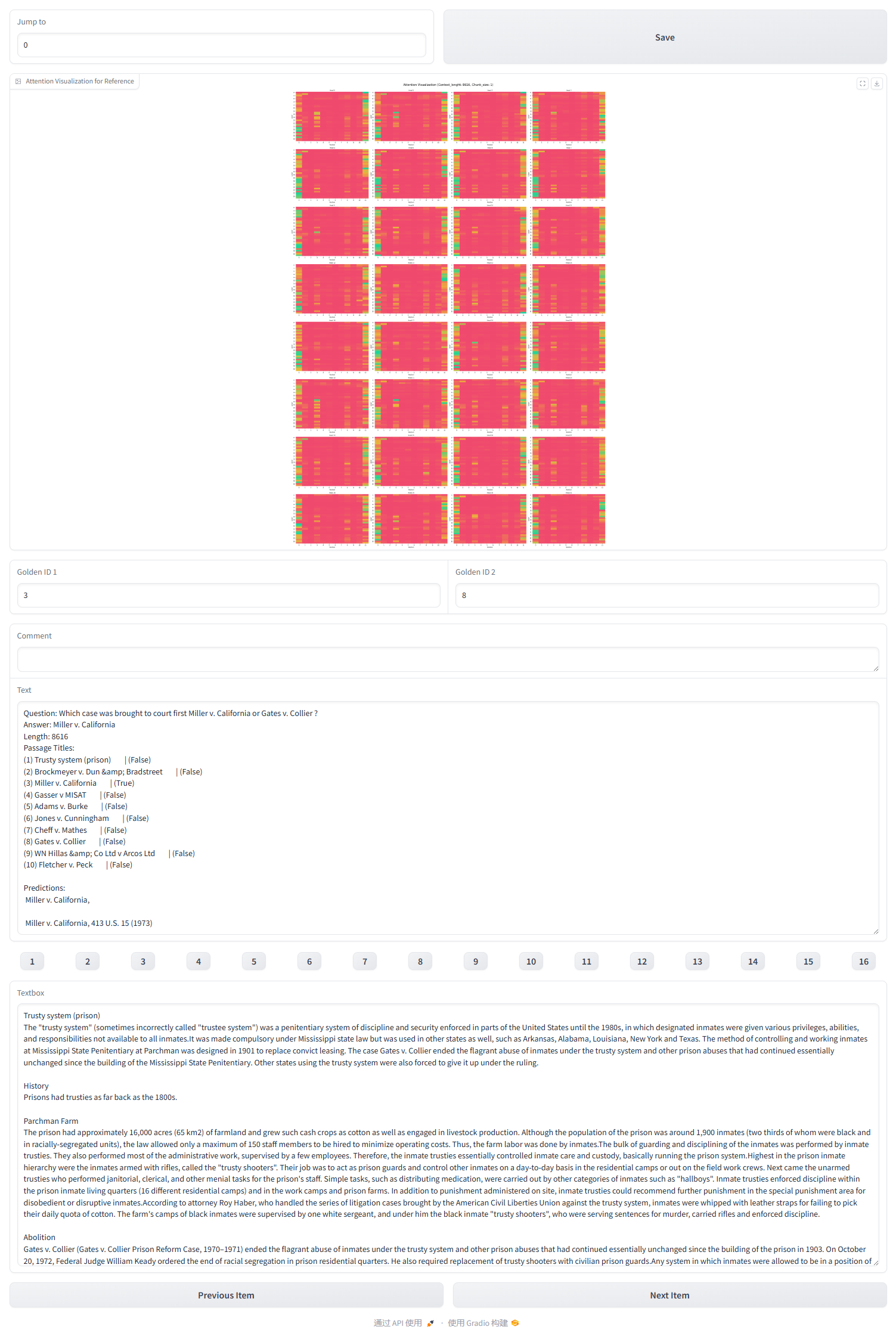}
    \caption{Our annotation interface with attention visualization for reference.}
    % , thus providing more accurate response.
    \label{fig:annotation}
\end{figure*}

Each question in the HotpotQA subset of LongBench has at least two relevant passages that contain essential information to answer the question. We manually reviewed each question and annotated its golden passages. Figure~\ref{fig:annotation} shows the annotation interface we used.

\subsection{Retrieval Score Calculation}
\label{app:retrieval_score}

Assume that for a given question $q$, we have:
\begin{itemize}
    \item A set of \emph{golden} (i.e., relevant) passages, $P_G$.
    \item A set of \emph{irrelevant} passages, $P_I$.
\end{itemize}
These passages are concatenated (in random order) to form the input context $\mathcal{C}$. An attention head $h$ assigns an attention weight $a_p$ to every passage $p \in \mathcal{C}$ (by summing the attention weights over tokens in the passage from the last token of the prompt). In the following we describe two retrieval metrics computed for head $h$.
\paragraph{EM Retrieval Score.} For the EM metric, the ranking of passages by their attention weights is used directly without considering the threshold $\epsilon$. Let
\[
X = |P_G|
\]
be the number of golden passages. Define a permutation $\sigma_h$ that sorts the passages in $\mathcal{C}$ in descending order of attention weight:
\[
a_{\sigma_h(1)} \ge a_{\sigma_h(2)} \ge \cdots \ge a_{\sigma_h(|\mathcal{C}|)}.
\]
Then, for a given question $q$, the EM Retrieval Score for head $h$ is defined as:
\begin{equation}
\small
\text{EM}_h(q) = 
\begin{cases}
1, & \text{if } \{\sigma_h(1), \sigma_h(2), \dots, \sigma_h(X)\} = P_G,\\[1mm]
0, & \text{otherwise.}
\end{cases}
\label{eq:em_single}
\end{equation}
That is, if the top $X$ passages (i.e., the $X$ passages with the highest attention weights) are exactly the golden passages, we consider the retrieval perfect and set $\text{EM}_h(q)=1$. Otherwise, $\text{EM}_h(q)=0$. The overall EM Retrieval Score for attention head $h$ is then the average over all test queries:
\begin{equation}
\mathcal{R}_h^{\text{EM}} = \frac{1}{|Q|}\sum_{q\in Q} \text{EM}_h(q),
\label{eq:em_overall}
\end{equation}
where $Q$ is the set of test questions.

\paragraph{F1 Retrieval Score.} For the F1 metric, we first use a fixed threshold $\epsilon$ to decide whether a passage is \emph{attended}. Specifically, define the set of passages attended by head $h$ for question $q$ as:
\begin{equation}
P_{A_h}(q) = \{\, p \in \mathcal{C} \mid a_p > \epsilon \,\}.
\label{eq:attended}
\end{equation}
Then, we compute the precision and recall based on the golden set $P_G$ and the attended set $P_{A_h}(q)$:
\begin{align}
\text{Precision} &= \frac{|P_G \cap P_{A_h}(q)|}{|P_{A_h}(q)|}, \label{eq:precision}\\[1mm]
\text{Recall} &= \frac{|P_G \cap P_{A_h}(q)|}{|P_G|}. \label{eq:recall}
\end{align}
The F1 Retrieval Score for head $h$ on question $q$ is then defined as the harmonic mean of precision and recall:
\begin{equation}
F1(q) = \frac{2 \cdot \text{Precision} \cdot \text{Recall}}{\text{Precision} + \text{Recall}},
\label{eq:f1_single}
\end{equation}
with the convention that if both precision and recall are zero, we set $F1(q)=0$. Finally, the overall F1 Retrieval Score for attention head $h$ is obtained by averaging over all test queries:
\begin{equation}
\mathcal{R}_h^{\text{F1}} = \frac{1}{|Q|}\sum_{q\in Q} F1(q).
\label{eq:f1_overall}
\end{equation}

Averaging these scores over the test set $Q$ yields $\mathcal{R}_h^{\text{EM}}$ [Eq.~\eqref{eq:em_overall}] and $\mathcal{R}_h^{\text{F1}}$ [Eq.~\eqref{eq:f1_overall}], which serve as our final metrics for evaluating the retrieval capability of attention head $h$.

% \paragraph{Calculation.}

% \paragraph{Analysis.}

% \section{Experimental Details}

% \subsection{Datasets}
% \label{app:datasets}

% \subsection{}
\section{More Implementation Details}
\label{app:imple}

We fine-tuned the Llama3.1-8B model with full parameters fine-tuning on 32~(2*16) 64G AMD INSTINCT MI250X GPUs. The distributed training was run with the DeepSpeed~\citep{rasley2020deepspeed} framework with ZeRO stage 2. The learning rate is 5e-6. We use AdamW~\citep{loshchilov2018decoupled} as the optimizer with $\beta_1=0.9$ and $\beta_2=0.999$. The template for the training sequence is: "Based on the following passages, answer the question.\textbackslash n\textbackslash n<passages>\textbackslash n\{body\}\textbackslash n\textbackslash n</passages>\textbackslash n\textbackslash n Question: \{question\}\textbackslash nAnswer: \{answer\}". The default EOS token is '</s>'.

\section{Analysis on Whole-layer Optimization}
\label{app:whole_layer}

We compared the retrieval scores of attention heads by layer between the \textit{Vanilla-SFT} baseline, \textit{MuDAF-strong} and \textit{MuDAF-weak}. Considering the different layers of selected attention heads, we can analyze the impact of a certain head on its layer.  Table~\ref{tab:layer_sta} shows the layer distribution of the two selection strategies. Considering some representative layers: 15, 16, 17, 25. For the layer 15, neither \textit{MuDAF-strong} nor \textit{MuDAF-weak} has heads from this layer, so most attention heads in this layer are not influenced with some of them being harmed actually (the line chart is below the zero threshold); For layer 16, both \textit{MuDAF-strong} and \textit{MuDAF-weak} select heads from it (four heads fror \textit{MuDAF-strong} and one head for \textit{MuDAF-weak}). As we can see, most attention heads are enhanced through \textit{MuDAF-strong}, even though they are not selected directly, and the improvement is much bigger than \textit{MuDAF-weak}. Meanwhile, nearly half of the attention heads also get enhanced through \textit{MuDAF-weak}, indicating that it is also helpful by selecting just one head; Finally, for layer 17 and 25, we can clearly observe that the retrieval scores of all attention heads are significantly improved when the corresponding strategy optimizes more attention heads at that layer~(i.e., \textit{MuDAF-strong} for layer 17 and \textit{MuDAF-weak} for layer 25). In the contrast, basically no improvement can be seen if no attention head is selected in that layer. We speculate that this phenomenon is related to the Grouped-Query Attention~(GQA)~\citep{ainslie2023gqa}, where \textit{query heads} are divided into several subgroups and each subgroup has only one corresponding key head. It also partly explains why \textit{MuDAF-weak} can still achieve appreciable improvements in the overall performance, given that there may exist some relatively strong attention heads within the same group.

~\begin{table}[h]
    \centering
    \begin{tabular}{ll}
    \toprule
    \textbf{Strategy}     & \textbf{Layer Distribution}  \\ \midrule
    \textit{MuDAF-strong} & \begin{tabular}[c]{@{}l@{}}Layer 13: 2 Heads\\ Layer 16: 3 Heads\\ Layer 17: 3 Heads\end{tabular}                                     \\ \midrule
    \textit{MuDAF-weak}   & \begin{tabular}[c]{@{}l@{}}Layer 16: 1 Head\\ Layer 19: 1 Head\\ Layer 25: 4 Heads\\ Layer 27: 1 Head\\ Layer 31: 1 Head\end{tabular} \\
    \bottomrule
    \end{tabular}
    \caption{The distribution of selected attention heads in two different selection strategies.}
    \label{tab:layer_sta}
\end{table}

\begin{figure*}[hp]
    \centering
    \includegraphics[width= 0.98\textwidth]{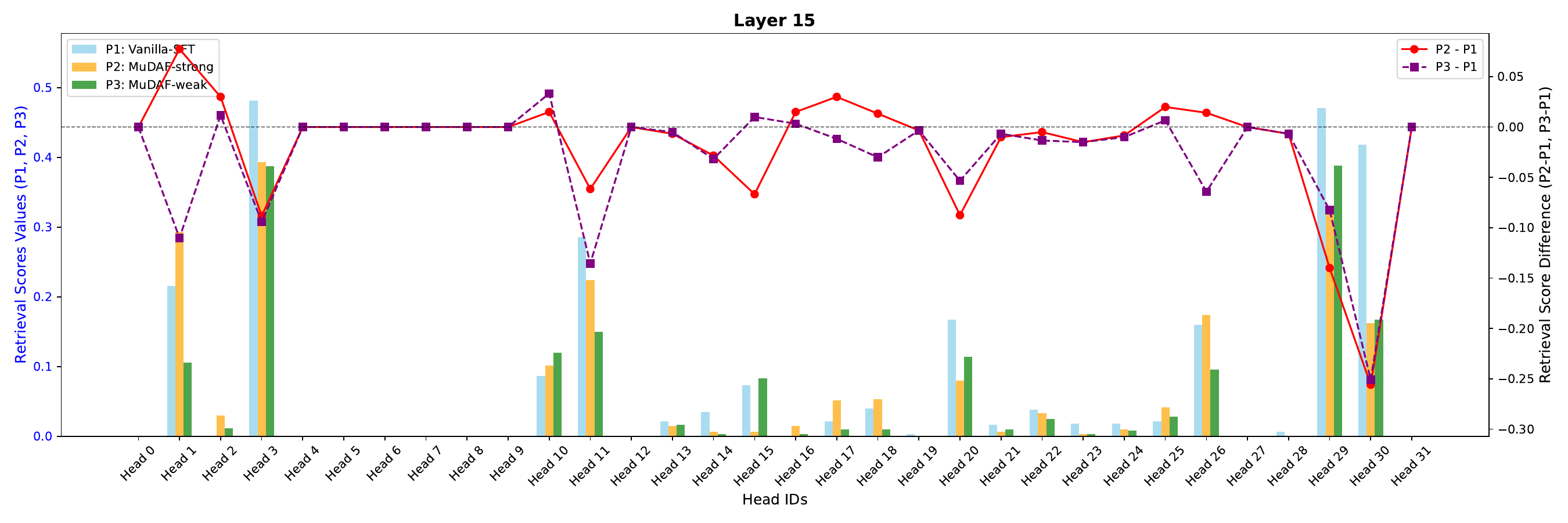}
    \includegraphics[width= 0.98\textwidth]{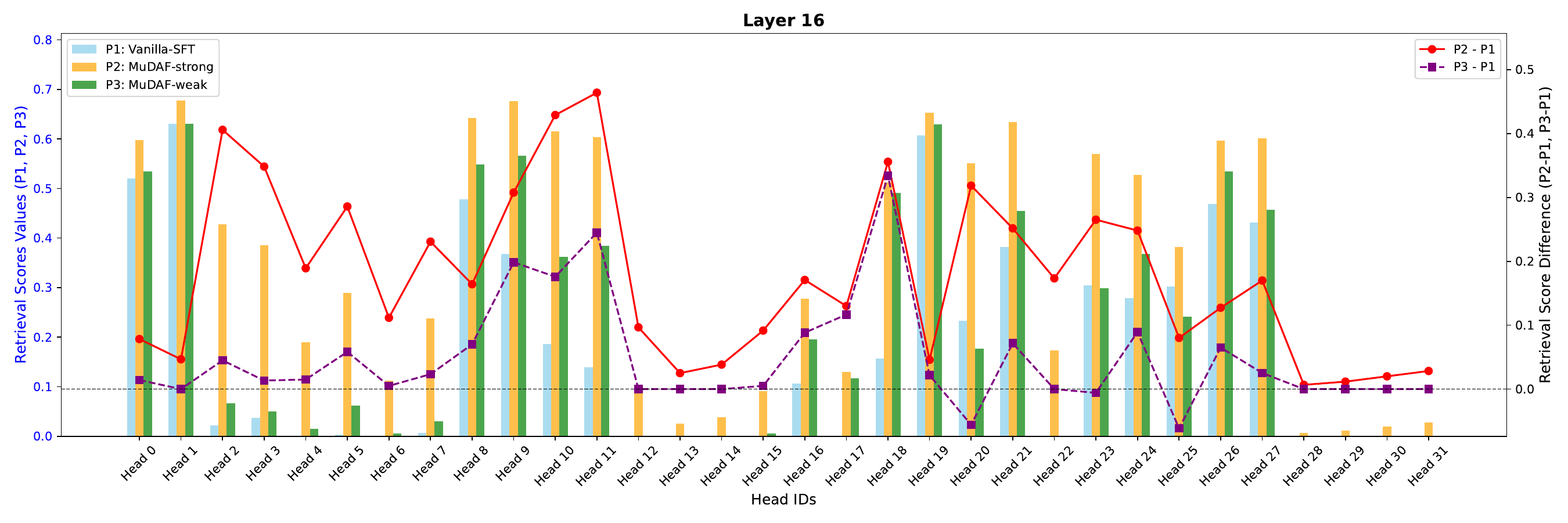}
    \includegraphics[width= 0.98\textwidth]{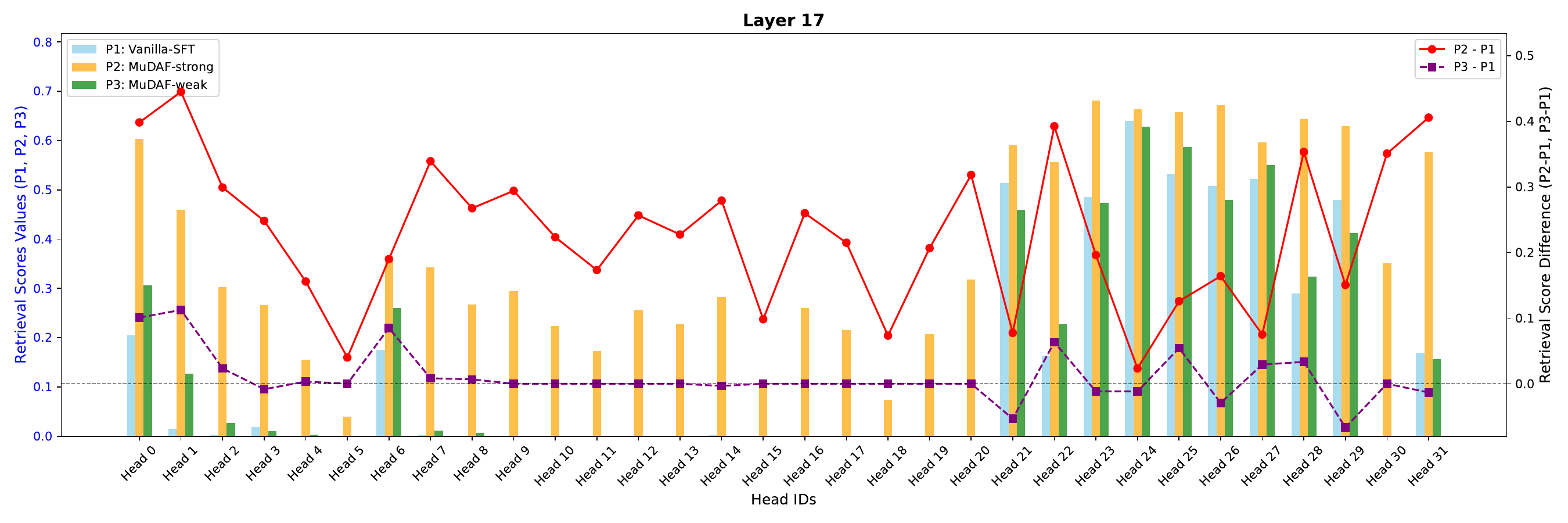}
    \includegraphics[width= 0.98\textwidth]{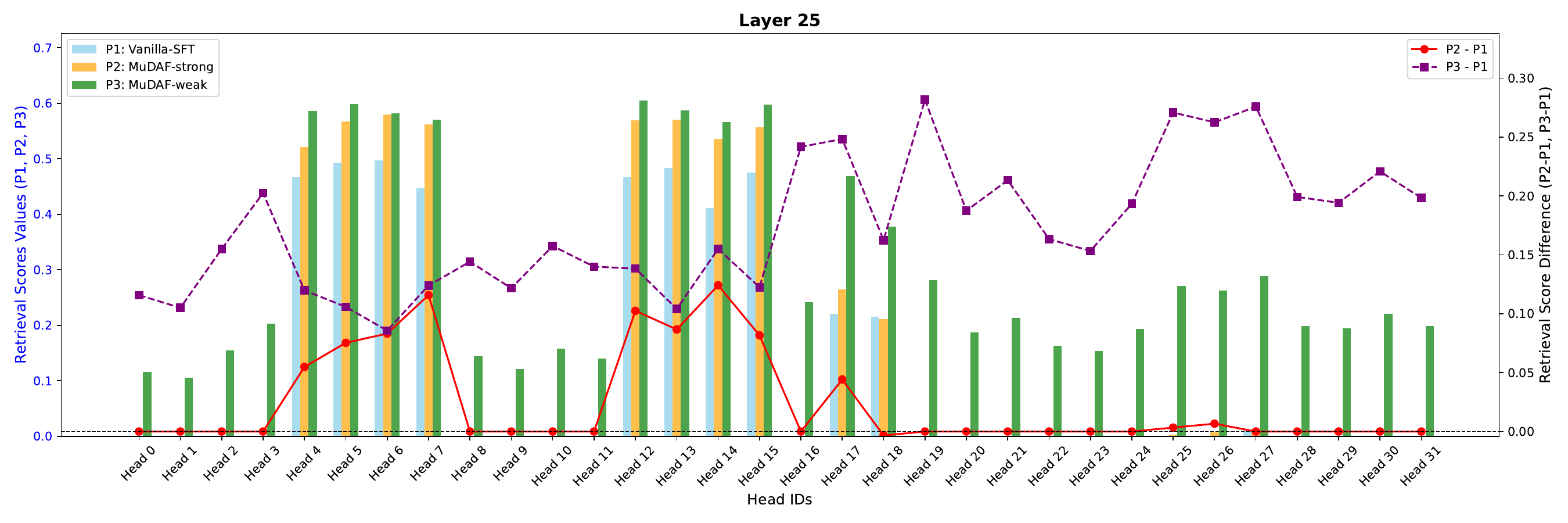}
    \caption{Retrieval scores of all attention heads listed by layer. The bars present the value of retrieval scores, while the line charts mean the difference between \textit{MuDAF-*} and the \textit{Vanilla-SFT} baseline.}
    % , thus providing more accurate response.
    \label{fig:layer_wise}
\end{figure*}

% \begin{figure*}[hp]
%     \centering
%     \includegraphics[width= 0.99\textwidth]{f1_17.pdf}
%     \includegraphics[width= 0.99\textwidth]{f1_25.pdf}
%     \caption{(Layer 17 and Layer 25) Retrieval scores of all attention heads listed by layer. The bars present the value of retrieval scores, while the line charts mean the different between \textit{MuDAF-*} and the \textit{Vanilla-SFT} baseline.}
%     % , thus providing more accurate response.
%     \label{fig:layer_wise}
% \end{figure*}

\section{More Experimental Results}
We provides full results for the ablation study on the number of selected attention heads in Table~\ref{tab:ablation}. 
% \subsection{Ablations}

% \begin{table*}[htbp]
% \centering
% \resizebox{0.98 \textwidth}{!}{
% \begin{tabular}{c|cccccccc}
% \toprule
% \multirow{2}{*}{\textbf{Strong or Weak}} & \multicolumn{4}{c}{\textbf{LongBench}}       & & \multicolumn{2}{c}{\textbf{ZeroSCROLLS}} & \multirow{2}{*}{\textbf{Avg.}} \\ \cmidrule{2-5} \cmidrule{7-8}
%                                          & HotpotQA & 2WikiMultihopQA & MuSiQue & Qasper & & MuSiQue             & Qasper             &                                \\ \midrule
% Strong Heads                             & \textbf{69.6}     & \textbf{66.2}            & \textbf{48.2}    & \textbf{40.0}   & & \textbf{31.2}                & \textbf{47.9 }              & \textbf{50.5}                           \\
% Weak Heads                               & 62.5     & 53.8            & 43.1    & 34.9   & & 23.2                & 41.8               &              43.2                 \\
% \bottomrule
% \end{tabular}
% }
% \caption{Ablations on whether to use strong attention heads or weak attention heads.}
% \end{table*}

\begin{table*}[htbp]
\centering
\resizebox{0.98 \textwidth}{!}{
\begin{tabular}{l|cccccccc}
\toprule
\multirow{2}{*}{\textbf{\begin{tabular}[c]{@{}c@{}}Num\\ Heads\end{tabular}}} & \multicolumn{4}{c}{\textbf{LongBench}}                          & & \multicolumn{2}{c}{\textbf{ZeroSCROLLS}} & \multirow{2}{*}{\textbf{Avg.}} \\ \cmidrule{2-5} \cmidrule{7-8}
                                                                              & HotpotQA      & 2WikiMultihopQA & MuSiQue       & Qasper      &  & MuSiQue             & Qasper             &                                \\ \midrule
n=0                                                                           &   46.8   &   50.5     &  28.9   & 29.2  & & 30.1                & 41.6               & 37.8                           \\ 
n=2                                                                           & 56.8          & 46.5            & 37.8          & 32.9         & & 26.8                & 42.6               & 40.5                           \\
n=4                                                                           & 64.7          & 57.9            & 41.7          & \textbf{36.9}& & 23.9                & 43.4               & 44.7                           \\
n=8                                                                           & 71.1          & 64.7            & 47.4          & 36.0         & & 35.2                & 42.4               & 49.4                           \\
n=16                                                                          & 72.2          & \textbf{68.8}   & \textbf{48.9} & 33.1         & & 28.3                & 40.2               & 48.6                           \\
n=32                                                                          & \textbf{72.6} & 66.1            & 48.3          & 33.2         & & \textbf{36.7}       & 40.8               & 49.6                           \\
n=64                                                                          & 69.4          & 65.8            & 48.7          & \textbf{36.9} & & 33.8                & \textbf{46.0}      & \textbf{50.0} \\
\bottomrule
\end{tabular}
}
\caption{Ablations on the number of selected attention heads.}
\label{tab:ablation}
\end{table*}

\section{Case Study}
\label{app:case_study}

\begin{figure*}
\includegraphics[width= \textwidth]{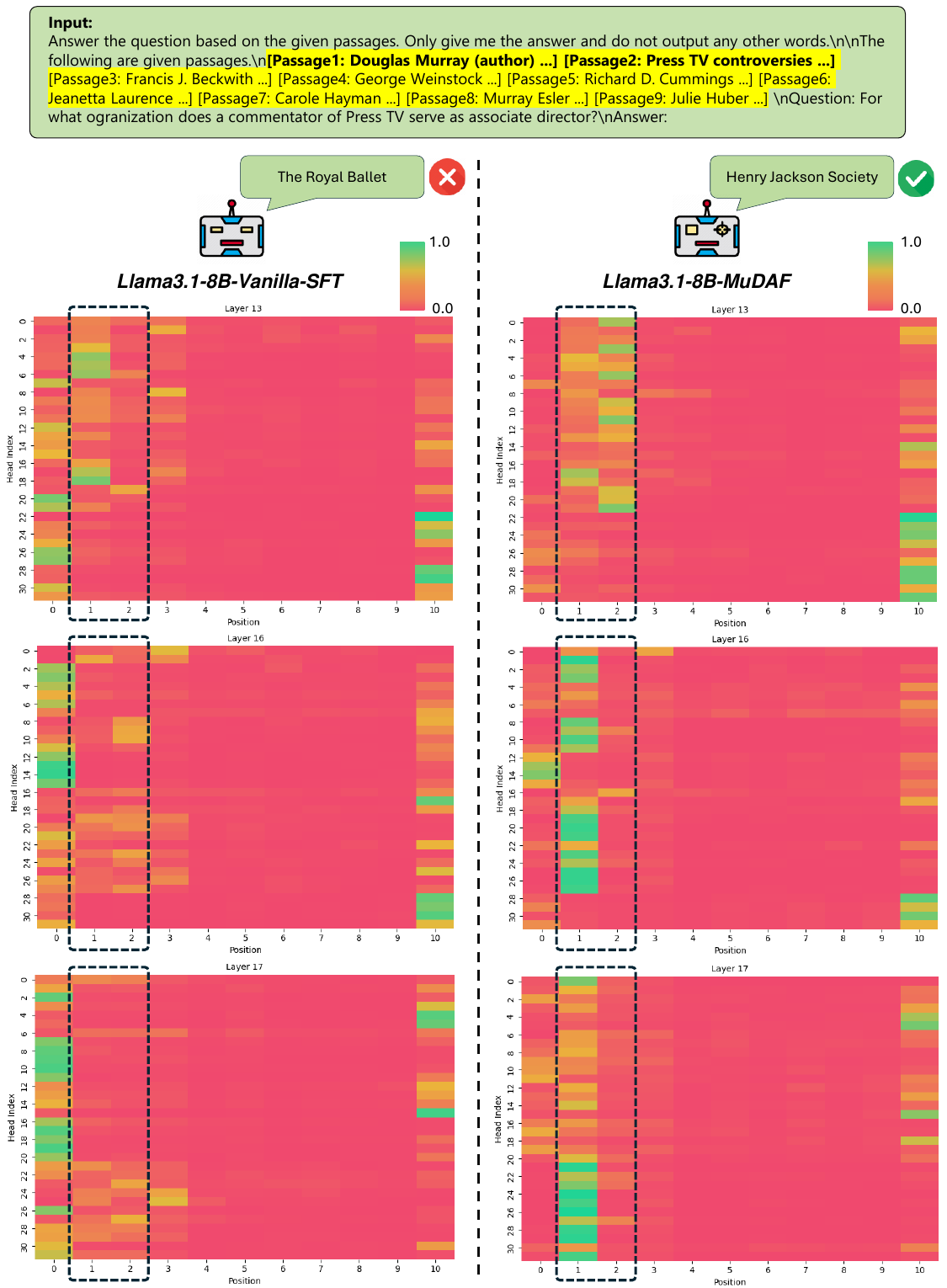}
\caption{Comparison of the output and passage-level attention distribution~(heatmaps below) in three different layers. This case contains 9 passages. The golden passages are passage\#1 and passage\#2~(in the dotted box). \textit{Llama3.1-8B-MuDAF} is more focused and can redirect the attention from the beginning part~(i.e., \#0) to the passages.}
\label{fig:case_study1}
\end{figure*}

\begin{figure*}
\includegraphics[width= \textwidth]{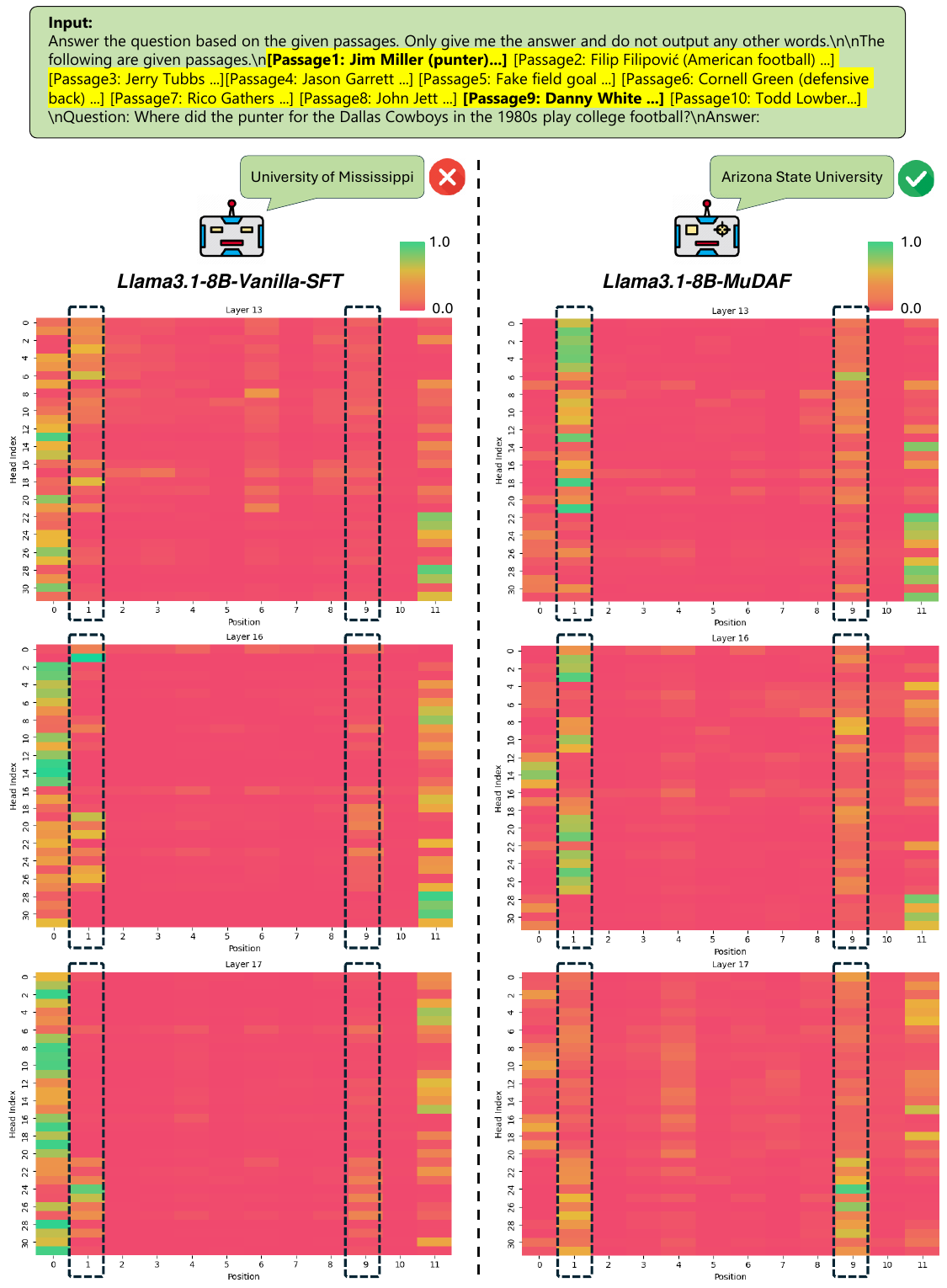}
\caption{This case contains 10 passages. The golden passages are passage\#1 and passage\#9~(in the dotted box).}
\label{fig:case_study2}
\end{figure*}

Figure~\ref{fig:case_study1} and Figure~\ref{fig:case_study2} show two cases that compare the output and attention distribution between Llama3.1-8B-\textit{Vanilla-SFT} and Llama3.1-8B-\textit{MuDAF}.~\method~effectively optimizes the attention distribution of the selected attention heads, making the model be more focused on relevant passages.

\end{document}